\documentclass[10pt,twocolumn,letterpaper]{article}

 \usepackage{cvpr}              
\usepackage{graphicx}
\usepackage{amsmath}
\usepackage{amssymb}
\usepackage{booktabs}
\usepackage{hyperref}
\usepackage{bm}
\usepackage{multirow}
\usepackage{makecell}

\hypersetup{colorlinks=true, urlcolor=red}
    
\def\paperID{13873} 
\def\confName{CVPR}
\def\confYear{2024}

\title{OSInsert: Towards High-authenticity and High-fidelity Image Composition}

\author{Jingyuan Wang, Li Niu \thanks{Corresponding author.} \\
MoE Key Lab of Artificial Intelligence, Shanghai Jiao Tong University\\
{\tt \{wjy0611,ustcnewly\}@sjtu.edu.cn} \\
}

\begin{document}
\maketitle
\definecolor{cvprblue}{rgb}{0.21,0.49,0.74}

\begin{abstract}
Generative image composition aims to regenerate the given foreground object in the background image to produce a realistic composite image. 
Some high-authenticity methods can  adjust foreground pose/view to be compatible with background, while some high-fidelity methods can preserve the foreground details accurately. However, existing methods can hardly achieve both goals at the same time. 
In this work, we propose a two-stage strategy to achieve both goals. In the first stage, we use high-authenticity method to generate reasonable foreground shape, serving as the condition of high-fidelity method in the second stage. 
The experiments on MureCOM dataset verify the effectiveness of our two-stage strategy. The code and model have been released at \href{https://github.com/bcmi/OSInsert-Image-Composition}{https://github.com/bcmi/OSInsert-Image-Composition} . 
\end{abstract}

\section{Introduction} \label{sec:intro}

Generative image composition (object insertion) is an important research direction at the intersection of computer vision and generative models, with broad practical applications in digital content creation, e-commerce product display, virtual reality, film and television post-production, and augmented reality~\cite{niu2021making}. For example, in e-commerce, merchants need to insert product images into different scene backgrounds to show product application effects without re-shooting; in film and television production, special effects artists need to seamlessly integrate virtual objects into real shot footage to create realistic visual effects. The core demand of all these applications is to generate composite images that are perceptually realistic—that is, the inserted foreground object must not only be consistent with the background in terms of spatial geometry, illumination, and perspective (authenticity) but also retain its own unique appearance details (fidelity) without distortion or loss.

With the rapid development of generative AI, especially diffusion models, a series of methods for generative image composition have been proposed~\cite{PBE,objectstitch,zhang2023controlcom,anydoor,song2025insert}. These methods have significantly improved the quality of composite images, but the inherent trade-off between authenticity and fidelity has not been effectively resolved, which has become a major bottleneck restricting the practical application of generative image composition technology. If the composite image lacks authenticity, the foreground object will appear ``out of place" relative to the background, and the copy-and-paste effect will be obvious. If the composite image lacks fidelity, the foreground object will lose its unique characteristics, and even the semantic information will be distorted (\emph{e.g.}, the color of a red car becomes blue, the texture of a wooden table disappears). Therefore, achieving both high authenticity and high fidelity is the key to breaking through the current technical limitations of generative image composition.
Existing generative image composition methods can be roughly categorized into high-authenticity methods~\cite{PBE,objectstitch}  and high-fidelity methods~\cite{zhang2023controlcom,anydoor,song2025insert} according to their core optimization objectives, and both types of methods have obvious inherent limitations.

High-authenticity methods (\emph{e.g.}, ObjectStitch~\cite{objectstitch}, Paint by Example~\cite{PBE}): This type of method takes the background compatibility as the core optimization goal, and its model design focuses on adjusting the pose, viewpoint, illumination, and scale of the foreground object to match the background. For example, ObjectStitch~\cite{objectstitch} uses a diffusion-based generative model to model the spatial and visual correlation between the foreground and the background, and can generate a foreground object with a reasonable perspective even when the reference foreground’s pose is very different from the expected pose. However, due to the strong transformation and generation capabilities of the model, high-authenticity methods often sacrifice the fine-grained details of the foreground object. For uncommonly seen objects (\emph{e.g.}, rare cultural relics, special industrial parts) or objects with complex surface details (\emph{e.g.}, patterned textiles, textured metal products), these methods will cause serious detail loss, color distortion, and texture blurring, which makes the generated foreground object lose its original semantic and visual characteristics.

High-fidelity methods (\emph{e.g.}, InsertAnything~\cite{song2025insert}, AnyDoor~\cite{anydoor}, ControlCom~\cite{zhang2023controlcom}): This type of method prioritizes the preservation of foreground appearance details, and its technical route is mainly based on in-context editing and detail-aware generation. For example, InsertAnything~\cite{song2025insert} concatenates the background image and the foreground reference image as the input of the diffusion model, providing sufficient in-context information for the model to learn the fine details of the foreground. This design enables high-fidelity methods to accurately retain the color, texture, shape, and local features of the foreground object, even for complex and rare objects. However, the drawback of this design is that the model lacks the ability to flexibly adjust the pose and viewpoint of the foreground object. When the expected pose/viewpoint of the foreground (determined by the background’s perspective and spatial layout) is dramatically different from that of the reference foreground, high-fidelity methods will directly copy the reference foreground into the background without effective transformation, resulting in an obvious copy-and-paste effect and serious lack of spatial compatibility.
In summary, the core problem of existing methods is that they cannot simultaneously model the spatial compatibility (authenticity) between the foreground and the background and the detail preservation (fidelity) of the foreground object. The root cause is that a single-stage generative model is difficult to balance these two conflicting optimization objectives in a single training and inference process: emphasizing authenticity will inevitably lead to over-transformation of the foreground and detail loss; emphasizing fidelity will restrict the model’s ability to adjust the foreground’s spatial attributes.

To solve the above problems, we propose a simple and effective two-stage framework OSInsert, which abandons the idea of a single-stage model to balance two conflicting objectives and instead uses a modular design to decouple the authenticity optimization and fidelity optimization into two independent stages. We first propose to decouple the high-authenticity and high-fidelity requirements of generative image composition into two sequential stages, which effectively avoids the trade-off between the two objectives in a single-stage model. The first stage is responsible for generating a spatially compatible foreground shape and pose (authenticity), and the second stage is responsible for filling the foreground region with fine-grained appearance details (fidelity). This decoupling design makes full use of the advantages of existing high-authenticity and high-fidelity methods without modifying their original model structures, which has the characteristics of simplicity and strong scalability.
Based on the two-stage decoupling strategy, we design the OSInsert framework by combining the high-authenticity method ObjectStitch and the high-fidelity method InsertAnything. We introduce the Segment Anything Model (SAM)~\cite{kirillov2023segment}  to extract the precise foreground mask from the intermediate result of the first stage, which serves as a key bridge between the two stages and ensures that the second stage only fills the foreground region without affecting the background content.

We conduct extensive quantitative and qualitative experiments on the MureCOM~\cite{lu2023dreamcom} dataset, and compare OSInsert with the state-of-the-art baselines and commercial models. The experimental results show that OSInsert not only generates a foreground object with a reasonable pose and viewpoint compatible with the background but also accurately preserves the fine details of the reference foreground, achieving a significant improvement over the baseline methods in both authenticity and fidelity. 

The rest of this paper is organized as follows: Section~\ref{sec:related} reviews the related work of generative image composition, including high-authenticity methods and high-fidelity methods. Section~\ref{sec:method} details the design of the OSInsert two-stage framework, including the specific steps of the first stage (authenticity generation) and the second stage (fidelity filling), as well as the role of the SAM model in mask extraction. Section~\ref{sec:exp} introduces the experimental setting and the experimental results. Section~\ref{sec:conclusion} concludes the full paper.

\section{Related Works} \label{sec:related}

In this section, we review the related work of generative image composition, and divide the existing methods into high-authenticity methods and high-fidelity methods according to their core optimization objectives.

\subsection{High-authenticity Image Composition}
High-authenticity methods focus on improving the spatial and visual compatibility between the foreground and the background, and their core idea is to model the contextual correlation between the background and the foreground, so that the generated foreground object can adapt to the pose, viewpoint, illumination, and scale of the background.
ObjectStitch~\cite{objectstitch} is a representative high-authenticity method, which is based on a diffusion model and designed for generative object compositing. ObjectStitch first erases the specified region of the background image to generate a masked background, and then feeds the masked background, the foreground reference image, and the bounding box into the diffusion model for generative inpainting. The model is trained to learn the spatial and visual correlation between the background and the foreground, and can generate a foreground object with a reasonable pose and viewpoint even when the reference foreground’s pose is inconsistent with the background. However, ObjectStitch has a serious problem of detail loss for complex objects, because the model tends to generate a foreground shape that is compatible with the background at the cost of sacrificing fine details.
Paint by Example~\cite{PBE} is another high-authenticity method based on diffusion models, which realizes exemplar-based image editing by learning the style and content mapping between the reference image and the target image. In the image composition task, Paint by Example can adjust the foreground’s illumination and texture to match the background, but it also has the problem of insufficient detail preservation for rare and complex objects.
The common characteristic of high-authenticity methods is that they use the background as the main context to guide the generation of the foreground, which enables the foreground to be well integrated into the background, but the over-reliance on the background context leads to the loss of the foreground’s own detail characteristics.

\subsection{High-fidelity Image Composition}
High-fidelity methods focus on preserving the fine-grained appearance details of the foreground object, and their core idea is to provide the model with sufficient in-context information of the foreground reference image, so that the model can accurately restore the color, texture, shape, and local features of the foreground.
InsertAnything~\cite{song2025insert} is a state-of-the-art high-fidelity method, which is based on diffusion models and in-context editing technology. InsertAnything concatenates the background image, the foreground reference image, and the bounding box as the input of the diffusion model, and the model uses the in-context information of the reference image to realize detail-aware object insertion. This design enables InsertAnything to accurately preserve the fine details of the foreground object, even for complex and rare objects. However, InsertAnything lacks the ability to adjust the pose and viewpoint of the foreground object, and will produce a serious copy-and-paste effect when the reference foreground’s pose is inconsistent with the background.
AnyDoor~\cite{anydoor} is a zero-shot object-level image customization method, which can insert different objects into the specified region of the background image while preserving the object’s details. ControlCom~\cite{zhang2023controlcom} is a controllable image composition method based on diffusion models, which uses control signals to guide the generation of the foreground, and has a certain ability to preserve details. However, both AnyDoor and ControlCom have limited ability to adjust the foreground’s pose and viewpoint, and cannot well solve the copy-and-paste problem in complex background scenarios.
The common characteristic of high-fidelity methods is that they take the foreground reference image as the main information source to guide the generation of the foreground, which enables the accurate preservation of foreground details, but the over-reliance on the reference image restricts the model’s ability to adapt the foreground to the background.

\begin{figure*}[t]
\centering
\includegraphics[width=0.8\linewidth]{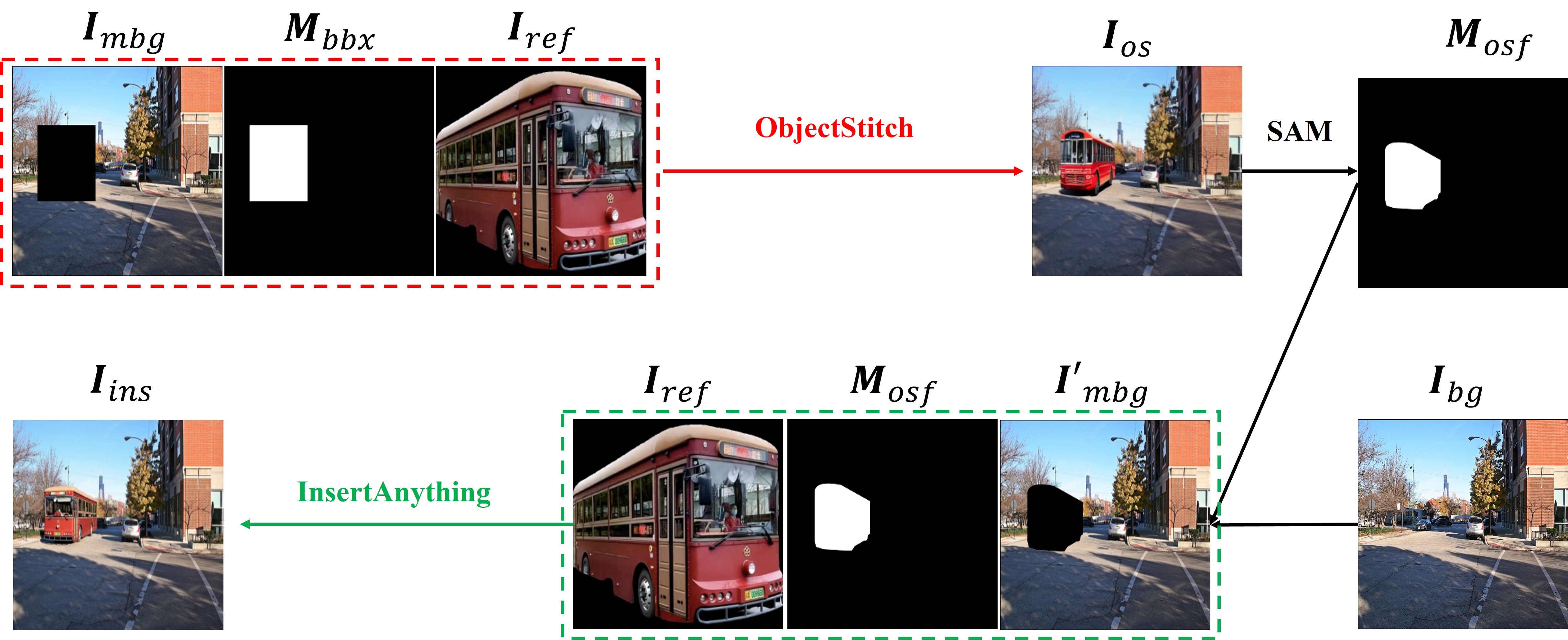} 
\caption {Illustration of our two-stage pipeline. In the first stage, we use  ObjectStitch~\cite{objectstitch} to generate the composite image with reasonable foreground pose/viewpoint and extract the foreground region. In the second stage, we use InsertAnything~\cite{song2025insert} to fill in the foreground region with the appearance details of reference image.}
\label{fig:pipeline}
\end{figure*}

\section{Method} \label{sec:method}

We propose a two-stage framework named OSInsert for generative image composition, which is specifically designed to address the long-standing trade-off between high authenticity and high fidelity in existing object insertion methods. The core design philosophy of OSInsert is to decouple the optimization of spatial compatibility (authenticity) and fine-grained detail preservation (fidelity) into two sequential and independent stages, where we leverage the respective strengths of state-of-the-art high-authenticity and high-fidelity methods. In the first stage, we adopt ObjectStitch \cite{objectstitch}, a representative high-authenticity generative compositing method, to generate an intermediate composite image with a foreground object that has a plausible pose, viewpoint, and geometric alignment with the background. We then use the Segment Anything Model (SAM) \cite{kirillov2023segment} to extract a high-precision foreground mask from this intermediate result, which serves as a critical spatial constraint for the second stage. In the second stage, we utilize InsertAnything \cite{song2025insert}, a cutting-edge high-fidelity in-context editing method, to fill the foreground region defined by the SAM-extracted mask with the original appearance details of the reference foreground object. This two-stage pipeline ensures that the foreground object is both seamlessly integrated into the background (authenticity) and retains its intrinsic visual characteristics (fidelity), and the overall workflow of OSInsert is illustrated in Figure \ref{fig:pipeline}.

We first formally define the input and output of the OSInsert framework to clarify the task setting:
\begin{itemize}
    \item \textbf{Input}: (1) A background image $\bm{I}_{bg}$ with a pre-specified axis-aligned bounding box $B$ that indicates the target placement of the foreground object; (2) A foreground reference image $\bm{I}_{ref}$ that contains the complete foreground object to be inserted, with all fine-grained visual details (e.g., color, texture, shape) preserved.
    \item \textbf{Output}: A final composite image $\bm{I}_{ins}$ where the foreground object is inserted into the background image $\bm{I}_{bg}$ at the position of bounding box $B$, with both high authenticity (pose/viewpoint/illumination compatibility with the background) and high fidelity (accurate preservation of the reference foreground's details).
\end{itemize}

The entire OSInsert pipeline is composed of three key components: the authenticity-oriented first stage based on ObjectStitch, the high-precision mask extraction module based on SAM, and the fidelity-oriented second stage based on InsertAnything. Each component is elaborated in detail in the following subsections, with mathematical formulations for all core operations to ensure reproducibility.

\subsection{The First Stage: Authenticity Generation}
The primary objective of the first stage is to generate a foreground object that is spatially and visually compatible with the background image, \emph{i.e.}, to achieve high authenticity. This stage eliminates the copy-and-paste effect that plagues high-fidelity methods by adjusting the foreground's pose, viewpoint, scale, and illumination to match the background context, which is realized by the generative inpainting capability of ObjectStitch. The first stage consists of two key steps: the construction of a masked background image and the generation of the intermediate composite image via ObjectStitch, both of which are described in detail below.

\subsubsection{Masked Background Image Construction}
Given the background image $\bm{I}_{bg}$ and the foreground placement bounding box $B$, we first convert the bounding box $B$ into a binary spatial mask $\bm{M}_{bbx} \in \{0,1\}^{H \times W}$, where $H$ and $W$ denote the height and width of the background image $\bm{I}_{bg}$, respectively. In the mask $\bm{M}_{bbx}$, the pixel values within the range of the bounding box $B$ are set to 1, and all other pixel values are set to 0. This binary mask $\bm{M}_{bbx}$ acts as a spatial indicator to mark the target region for foreground insertion in the background image.

We then erase the pixel content of the background image $\bm{I}_{bg}$ within the region defined by $\bm{M}_{bbx}$ to construct the masked background image $\bm{I}_{mbg}$. The erasure operation is implemented by setting all pixel values in the masked region to $0$ (black), a standard preprocessing step for generative inpainting tasks that guides the model to generate new content in the erased region instead of retaining the original background pixels. Mathematically, the construction of the masked background image $\bm{I}_{mbg}$ is formulated as:
\begin{equation}
I_{mbg}(x,y) =
\begin{cases}
0, & (x,y) \in \bm{M}_{bbx}, \\
I_{bg}(x,y), & (x,y) \notin \bm{M}_{bbx},
\end{cases}
\end{equation}
where $(x,y)$ represents the 2D pixel coordinates of the image at row $x$ and column $y$. The masked background image $\bm{I}_{mbg}$ retains all the original visual information of the background outside the bounding box $B$, while the target insertion region is blank, providing a clean canvas for ObjectStitch to generate a background-compatible foreground object.

\subsubsection{Intermediate Composite Image Generation}
We feed the three core inputs (the masked background image $\bm{I}_{mbg}$, the bounding box mask $\bm{M}_{bbx}$, and the foreground reference image $\bm{I}_{ref}$) into the pre-trained ObjectStitch~\cite{objectstitch} model for generative inpainting. ObjectStitch is a diffusion model-based generative object compositing method that is trained to model the spatial and visual correlations between the background and foreground. It can generate a foreground object with a reasonable pose and viewpoint that aligns with the background context, even when the pose/viewpoint of the reference foreground $\bm{I}_{ref}$ is drastically different from the expected pose/viewpoint in the background $\bm{I}_{bg}$.

The generative process of ObjectStitch for producing the intermediate composite image $\bm{I}_{os}$ is simplified as the following functional mapping:
\begin{equation}
\bm{I}_{os} = \mathcal{F}_{OS}(\bm{I}_{mbg}, \bm{M}_{bbx}, \bm{I}_{ref}),
\end{equation}
where $\mathcal{F}_{OS}$ denotes the generative function of the pre-trained ObjectStitch model, which encapsulates all the diffusion-based inpainting and compositing capabilities of the model. The output of this step, the intermediate composite image $\bm{I}_{os}$, has a foreground object that is perfectly aligned with the background in terms of pose, viewpoint, scale, and illumination, thus achieving high authenticity. However, as an inherent limitation of high-authenticity methods, the foreground object in $\bm{I}_{os}$ suffers from noticeable detail loss such as blurred textures, distorted colors, and missing fine features, especially for objects with complex surface details or rare visual characteristics. This detail loss is not a flaw of the first stage but a necessary trade-off for achieving background compatibility, and the lost details will be fully recovered in the second stage of the OSInsert pipeline.

\subsubsection{Foreground Mask Extraction}
After generating the intermediate composite image $\bm{I}_{os}$ with a background-compatible foreground, we need to extract a high-precision foreground mask to define the exact spatial range of the generated foreground object. This mask is a critical bridge connecting the first and second stages, as it restricts the high-fidelity filling operation in the second stage to the foreground region only, preventing the foreground details from "bleeding" into the background and ensuring the integrity of the background image. We use SAM~\cite{kirillov2023segment}, a foundation model for universal image segmentation with strong zero-shot capability, to perform this mask extraction task.

SAM takes an image and a prompt as input and outputs a high-precision binary mask for the target object in the image. In our pipeline, we use the original foreground placement bounding box $B$ as the prompt for SAM, and input the intermediate composite image $\bm{I}_{os}$ and the bounding box $B$ into the pre-trained SAM model. SAM segments the foreground object from the background in $\bm{I}_{os}$ with pixel-level precision, generating the high-precision foreground mask $\bm{M}_{osf} \in \{0,1\}^{H \times W}$, where the pixel values corresponding to the foreground object are set to 1 and the background pixel values are set to 0. The mask extraction process is formulated as:
\begin{equation}
\bm{M}_{osf} = \mathcal{F}_{SAM}(\bm{I}_{os}, B),
\end{equation}
where $\mathcal{F}_{SAM}$ represents the segmentation function of the pre-trained SAM model.

Compared with the original bounding box mask $\bm{M}_{bbx}$, the SAM-extracted foreground mask $\bm{M}_{osf}$ has two key advantages: first, $\bm{M}_{osf}$ is a pixel-level mask that perfectly fits the contour of the foreground object in $\bm{I}_{os}$, rather than a rigid rectangular mask; second, $\bm{M}_{osf}$ accurately excludes the background pixels within the bounding box $B$, avoiding the inclusion of irrelevant background regions in the foreground mask. These advantages ensure that the high-fidelity filling in the second stage is performed only on the actual foreground region, which is essential for generating a realistic composite image.

Finally, we construct a new masked background image $\bm{I}_{mbg}'$ using the original background image $\bm{I}_{bg}$ and the high-precision foreground mask $\bm{M}_{osf}$. The construction method of $\bm{I}_{mbg}'$ is consistent with that of $\bm{I}_{mbg}$: we erase the pixel content of $I_{bg}$ within the region defined by $\bm{M}_{osf}$ (set to 0), as shown in the following formulation:
\begin{equation}
I_{mbg}'(x,y) =
\begin{cases}
0, & (x,y) \in \bm{M}_{osf}, \\
I_{bg}(x,y), & (x,y) \notin \bm{M}_{osf}.
\end{cases}
\end{equation}
The new masked background image $\bm{I}_{mbg}'$ is the key input of the second stage, which retains the complete original background and marks the exact foreground region (with pixel-level precision) that needs to be filled with high-fidelity details from the reference foreground $\bm{I}_{ref}$.

\subsection{The Second Stage: Fidelity Filling}
The second stage of the OSInsert pipeline is dedicated to recovering the fine-grained visual details of the reference foreground object and filling them into the foreground region defined by the mask $\bm{M}_{osf}$ in the new masked background image $\bm{I}_{mbg}'$, \emph{i.e.}, to achieve high fidelity. The core constraint of this stage is to retain the background-compatible pose, viewpoint, and contour of the foreground object generated in the first stage, while only replacing the appearance details of the foreground with the original details from the reference image $\bm{I}_{ref}$. This constraint is naturally satisfied by using the SAM-extracted mask $\bm{M}_{osf}$ as a spatial guide, and we leverage InsertAnything~\cite{song2025insert}, a high-fidelity in-context editing method, to perform the detail filling operation.

InsertAnything is a Diffusion Transformer (DiT) based image insertion method that adopts an in-context editing strategy. It concatenates the background image and the foreground reference image as input, which provides the model with sufficient in-context information about the foreground's appearance details, enabling it to preserve the color, texture, shape, and all fine-grained features of the reference foreground with high accuracy. Unlike high-authenticity methods that modify the foreground's spatial attributes to match the background, InsertAnything focuses on detail preservation and does not alter the spatial shape of the target region, which makes it the ideal method for the second stage of our pipeline.

In the second stage, we feed three core inputs into the pre-trained InsertAnything model: the new masked background image $\bm{I}_{mbg}'$, the high-precision foreground mask $\bm{M}_{osf}$, and the original foreground reference image $\bm{I}_{ref}$. The mask $\bm{M}_{osf}$ acts as a strict spatial constraint that guides InsertAnything to only fill the erased foreground region in $\bm{I}_{mbg}'$ and not modify any other parts of the background image. InsertAnything then fills the masked foreground region with the fine-grained appearance details from the reference image $\bm{I}_{ref}$, while strictly following the contour and spatial shape of the foreground defined by $\bm{M}_{osf}$ (\emph{i.e.}, the background-compatible pose/viewpoint generated in the first stage).

The generation process of the final high-quality composite image $\bm{I}_{ins}$ via InsertAnything is formulated as the following functional mapping:
\begin{equation}
\bm{I}_{ins} = \mathcal{F}_{IA}(\bm{I}_{mbg}', \bm{M}_{osf}, \bm{I}_{ref}),
\end{equation}
where $\mathcal{F}_{IA}$ denotes the generative function of the pre-trained InsertAnything model, which encapsulates its in-context editing and detail-preserving capabilities.

The key advantage of this two-stage decoupling design is that it completely avoids the trade-off between authenticity and fidelity that plagues single-stage methods. In the first stage, we focus solely on optimizing authenticity without considering detail preservation, and in the second stage, we focus solely on optimizing fidelity without modifying the spatial attributes of the foreground. This design allows us to fully leverage the strengths of ObjectStitch (high authenticity) and InsertAnything (high fidelity) while avoiding their respective weaknesses. The final composite image $\bm{I}_{ins}$ generated by the second stage thus integrates the best of both worlds: the foreground object has a plausible pose, viewpoint, and illumination that are perfectly compatible with the background (from the first stage), and simultaneously retains all the fine-grained visual details of the reference foreground object (from the second stage). This effectively solves the core dilemma of existing generative image composition methods and achieves both high authenticity and high fidelity for the first time.

\begin{figure*}[t]
\centering
\includegraphics[width=1.0\linewidth]{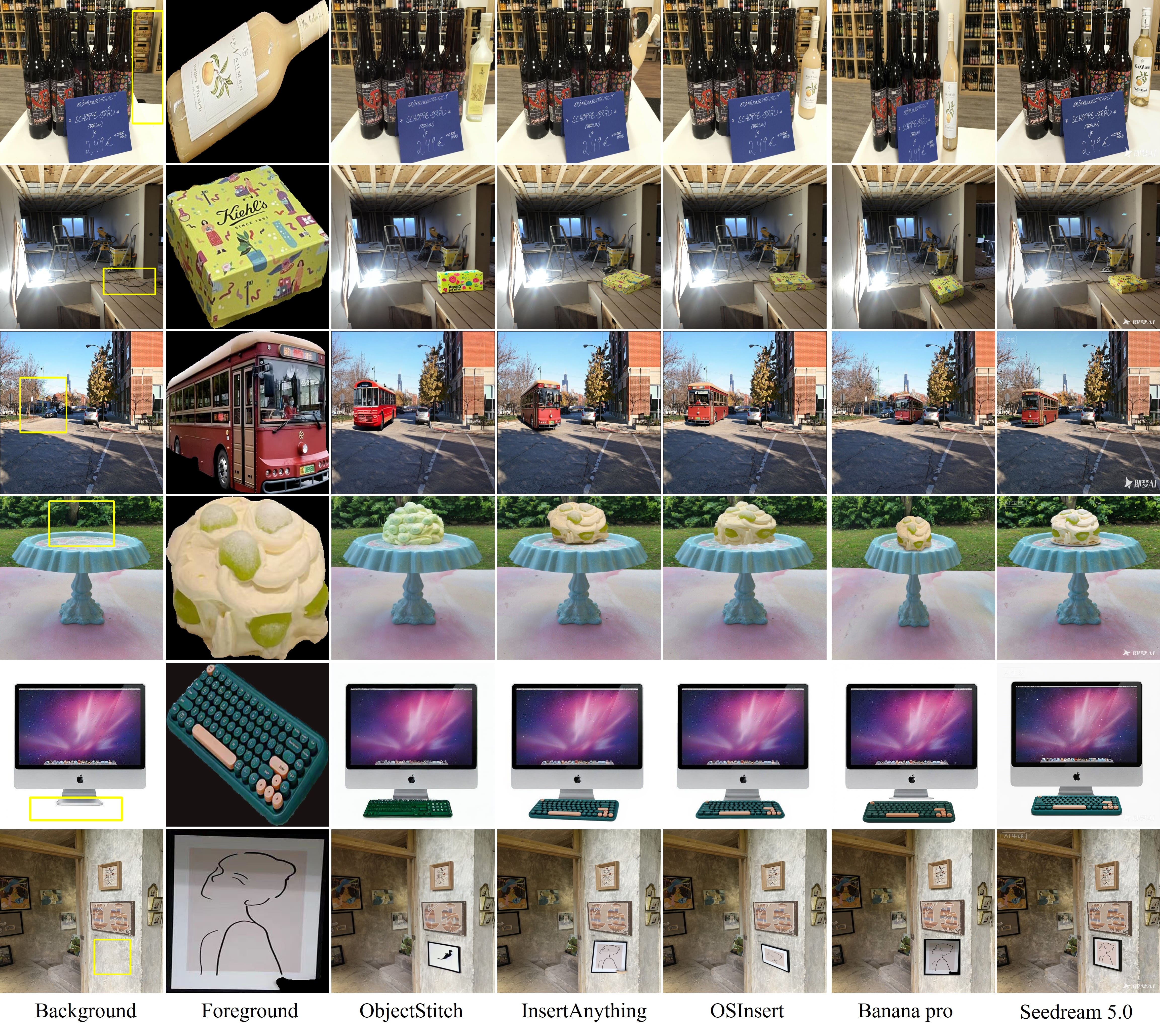} 
\caption{The visualization results of different generative composition methods on MureCom~\cite{lu2023dreamcom} dataset.
    From left to right in each row, we show the background with foreground bounding box, five reference images of the same foreground object,  the generated results of ObjectStitch~
    \cite{objectstitch}, InsertAnything~\cite{song2025insert}, our OSInsert, Banana pro~\cite{team2024gemini}, and Seedream 5.0~\cite{seedream2025seedream}.}
\label{fig:results}
\end{figure*}

\section{Experiment} \label{sec:exp}

In this section, we will introduce the experimental setup of our work, including the dataset used in the experiments and the baseline methods for comparison. Then, we will present and discuss the experimental results. 

\subsection{Dataset}
We conduct all experiments on the MureCOM dataset~\cite{lu2023dreamcom}, which is a dedicated benchmark dataset for generative image composition tasks. The MureCOM dataset contains a large number of background images, foreground reference images, and corresponding bounding box annotations for foreground placement. The main characteristics of the MureCOM dataset are as follows: 1) Rich scene diversity: The dataset contains a variety of background scenes, including indoor scenes (\emph{e.g.}, living room, bedroom, office), outdoor scenes (\emph{e.g.}, street, park), and natural scenes (\emph{e.g.}, lake, forest), which can fully test the model’s adaptability to different background contexts.
2) Complex foreground objects: The foreground objects in the dataset cover a variety of categories, including daily necessities (\emph{e.g.}, cup), transportation tools (\emph{e.g.}, car, airplane), animals (\emph{e.g.}, cat, dog), and industrial products (\emph{e.g.}, computer, mobile phone). Among them, many foreground objects have complex surface textures and fine details (\emph{e.g.}, patterned ceramic cups, textured leather sofas), which can effectively test the model’s detail preservation ability.
3) Diverse pose/viewpoint differences: The dataset is designed to have obvious pose/viewpoint differences between the foreground reference image and the expected foreground in the background, which can fully test the model’s ability to adjust the foreground pose/viewpoint to match the background (authenticity).
4) Standard annotations: Each sample in the dataset contains a background image, a foreground reference image, and a precise bounding box annotation for the foreground placement, which provides a unified input standard for the image composition model.

\subsection{Baseline Methods}
To comprehensively validate the performance of our proposed OSInsert framework and demonstrate its superiority in balancing high authenticity and high fidelity for generative image composition, we conduct extensive comparative experiments against a diverse set of baseline methods, covering both state-of-the-art open-source academic models and representative close-source commercial models with advanced image generation capabilities. Specifically, we compare OSInsert with two core open-source baselines that form the foundation of our two-stage design: ObjectStitch~\cite{objectstitch}, a classic high-authenticity generative object compositing method renowned for its ability to generate background-compatible foreground pose and viewpoint; and InsertAnything~\cite{song2025insert}, a cutting-edge high-fidelity in-context editing method that excels at precise preservation of foreground appearance details. Beyond academic baselines, we further extend our comparisons to two well-recognized close-source commercial models, Banana pro~\cite{team2024gemini} and Seedream 5.0~\cite{seedream2025seedream}, which are widely adopted in real-world image editing and composition tasks and represent the current industrial level of generative image technology. 
 
\subsection{Experimental Results}

From Fig.~\ref{fig:results}, we can clearly observe the performance characteristics of all compared methods in the generative image composition task, with distinct strengths and weaknesses emerging across open-source academic baselines and close-source commercial models. Specifically, ObjectStitch~\cite{objectstitch}, as a representative high-authenticity academic method, demonstrates a strong capability to generate foreground objects with viewpoint and pose that are naturally compatible with the background spatial context and geometric layout; however, it suffers from notable deficiencies in preserving the fine-grained appearance details of the original foreground reference, often leading to distorted colors, blurred textures, and lost local features, especially for objects with complex surface details or unique visual characteristics. In contrast, InsertAnything~\cite{song2025insert}, a state-of-the-art high-fidelity academic method, excels at the precise preservation of the foreground’s intrinsic appearance details, retaining the original color, texture, and shape of the reference foreground with high fidelity; yet it exhibits a critical limitation in spatial adaptation, struggling to adjust the foreground’s viewpoint and pose to match the background when there is a significant mismatch between the reference foreground’s spatial attributes and the expected ones in the background, thus resulting in an obvious copy-and-paste effect that undermines the overall realism of the composite image. Our proposed OSInsert effectively mitigates the aforementioned drawbacks of the two single-purpose academic methods by virtue of its two-stage decoupling design, seamlessly combining the strong background-compatible viewpoint/pose generation of ObjectStitch~\cite{objectstitch} and the high-precision detail preservation of InsertAnything~\cite{song2025insert}. As a result, OSInsert achieves simultaneous excellence in both fine-grained appearance detail retention and adaptive foreground viewpoint/pose adjustment, generating composite images where the foreground is not only spatially aligned with the background but also visually consistent with the reference object, without any unintended alterations to the original background. 

For the competitive close-source commercial models, Banana pro and Seedream 5.0 show impressive overall performance in balancing basic viewpoint/pose rationality and foreground detail faithfulness, outperforming single academic baselines in certain general scene compositions thanks to their large-scale pre-training on diverse visual data. Nevertheless, these commercial models present prominent practical flaws: on the one hand, the generated foreground objects fail to achieve accurate spatial alignment with the pre-provided bounding boxes for foreground placement, with the inserted objects often being slightly offset, scaled improperly, or partially out of the designated bounding box region; on the other hand, the original background’s color tone and luminance present slight yet discernible alterations in the generated composite images, which disrupts the visual consistency and integrity of the background scene. This dual issue of spatial misalignment and background tone distortion severely restricts their applicability in practical image composition tasks that require precise spatial placement and unaltered background visual characteristics, while our OSInsert maintains strict adherence to the given bounding box constraints throughout the two-stage generation process and preserves the original background’s color tone, luminance and all visual details intact, ensuring both spatial precision and holistic visual quality of the composite results.

\section{Conclusion} \label{sec:conclusion}
This work proposes OSInsert, a two-stage framework to resolve the authenticity-fidelity trade-off in generative image composition. We first use ObjectStitch to generate a background-compatible foreground and SAM to extract its precise mask, then employ InsertAnything to fill the mask region with fine-grained foreground details from the reference image. Experiments on the MureCOM dataset confirm OSInsert outperforms single-stage baselines, achieving both realistic foreground-background integration and accurate detail preservation. As a simple, effective baseline, OSInsert offers valuable insights for generative image composition research, with open-sourced code and models supporting further exploration in the field.

{
    \small
    \bibliographystyle{ieeenat_fullname}
    \bibliography{main.bbl}
}


\end{document}